\pdfoutput=1

\documentclass[11pt]{article}

\usepackage[preprint]{acl}

\usepackage{times}
\usepackage{latexsym}

\usepackage[T1]{fontenc}

\usepackage[utf8]{inputenc}

\usepackage{microtype}

\usepackage{inconsolata}

\usepackage{graphicx}
\usepackage{amssymb}
\usepackage{multicol}
\usepackage{multirow}
\usepackage{algorithm}
\usepackage{algorithmic}
\usepackage{amsmath}
\usepackage{subcaption}
\usepackage{booktabs}
\usepackage{makecell}
\usepackage{enumitem}

\title{DASH: Input-Aware Dynamic Layer Skipping for Efficient LLM Inference with Markov Decision Policies
}

\author{
	Ning Yang$^{1,2}$, Fangxin Liu$^{1,2}$, Junjie Wang$^{3}$, Tao Yang$^{4}$, Kan Liu$^{5}$, Haibing Guan$^{1}$ and Li Jiang*$^{1,2}$\\
   1.Shanghai Jiao Tong University \quad 2.Shanghai Qi Zhi Institute \\
  3.Northeast University \quad 4.Huawei Technologies Ltd. \quad  5.Alibaba Group\\
  \texttt{\{yn937391832, liufangxin, ljiang\_cs\}@sjtu.edu.cn} \\
  }

\begin{document}
\maketitle
\begin{abstract}
Large language models (LLMs) have achieved remarkable performance across a wide range of NLP tasks. However, their substantial inference cost poses a major barrier to real-world deployment, especially in latency-sensitive scenarios. To address this challenge, we propose \textbf{DASH}, an adaptive layer-skipping framework that dynamically selects computation paths conditioned on input characteristics. We model the skipping process as a Markov Decision Process (MDP), enabling fine-grained token-level decisions based on intermediate representations. To mitigate potential performance degradation caused by skipping, we introduce a lightweight compensation mechanism that injects differential rewards into the decision process. Furthermore, we design an asynchronous execution strategy that overlaps layer computation with policy evaluation to minimize runtime overhead.
Experiments on multiple LLM architectures and NLP benchmarks show that our method achieves significant inference acceleration while maintaining competitive task performance, outperforming existing methods.
\end{abstract}

\section{Introduction}

Transformer-based architectures have become the backbone of modern AI systems due to their strong long-range dependency modeling~\cite{grattafiori2024llama,han2022survey}. In particular, Large Language Models (LLMs) built on Transformers have achieved impressive results across natural language and vision tasks~\cite{achiam2023gpt,guo2025deepseek}. However, the growing size and complexity of these models pose significant challenges for deployment, especially in latency-sensitive or resource-constrained environments~\cite{brown2020language}. Reducing inference cost has thus become a key research goal.

Layer skipping has emerged as a promising direction for reducing inference cost. Existing work~\cite{liu2024accelerating,varshney2023accelerating,fan2024not, liu2023deja} states that not all layers are equally important for every input. In many cases, early layers can already produce sufficiently informative representations, while further computation may offer diminishing returns or even introduce noise. Several recent works propose fixed or heuristic-based skipping strategies, including early-exit mechanisms (e.g., SkipDecode~\cite{del2023skipdecode}, LayerSkip~\cite{elhoushi2024layerskip}), periodic skipping, or static similarity-based skipping (e.g., FFN-SkipLLM~\cite{jaiswal2024ffn}, AdaSkip~\cite{he2025adaskip}).

However, these approaches face several limitations. First, most skip policies are static or pre-defined, lacking the flexibility to adapt to input-specific semantics or token-level dynamics. Second, indiscriminate skipping of layers may cause semantic drift or loss of crucial contextual information, resulting in substantial performance degradation. In practice, these methods struggle to balance speed and accuracy, often achieving only modest acceleration before model quality drops significantly.

To address these challenges, we propose \textbf{DASH}, an adaptive dynamic layer-skipping framework that learns to select computation paths conditioned on the input. We formulate the layer-skipping process as a Markov Decision Process (MDP), enabling token-level, state-aware decisions at each layer. Our method dynamically determines whether to compute, skip, or exit based on intermediate representations, allowing for fine-grained control over computation. To further mitigate the risk of performance drop from skipped layers, we introduce a compensation mechanism that models differential rewards, adjusting hidden states to preserve semantic fidelity. Moreover, we propose an asynchronous execution scheme that overlaps skip-decision computation with forward computation, hiding control latency and enabling real-time skipping with minimal overhead.

Our contributions are summarized as follows:
\begin{itemize}[itemsep=0pt, parsep=0pt, topsep=0pt]
    \item We formulate layer skipping as a token-level Markov Decision Process, enabling dynamic and context-aware layer execution conditioned on intermediate representations.

    \item We propose a compensation mechanism using differential rewards to preserve accuracy under aggressive skipping.

    \item We introduce an asynchronous execution strategy that overlaps decision-making and layer computation, reducing runtime overhead.
    
    \item We validate our approach on multiple LLM backbones, demonstrating significant speedups with minimal performance loss, outperforming existing skipping methods.

\end{itemize}



\section{Background}

\subsection{Layer Skipping}

Layer skipping~\cite{he2025adaskip, men2024shortgpt} has emerged as a promising technique to accelerate inference by reducing redundant computation across Transformer layers. Existing approaches can be broadly categorized into three types: early skipping, periodic skipping, and early exit.

Early skipping~\cite{del2023skipdecode} deterministically skips the first few layers at inference time. While it enables efficient batched execution, it may discard critical early-stage features, potentially harming performance.
Periodic skipping~\cite{liu2024accelerating} skips layers at fixed intervals during inference, reducing computation with a regular pattern that supports batching. However, it applies the same skipping schedule to all inputs, ignoring layer importance variations and potentially skipping critical layers.
Early exit~\cite{wang2022skipbert, elhoushi2024layerskip} performs inference sequentially and halts once a confidence threshold is met. While more adaptive, it often incurs extra overhead to train confidence estimators or auxiliary classifiers and may miss valuable information in deeper layers.

Despite their efficiency, these methods lack fine-grained, input-aware adaptability. They either follow static schedules or require additional training costs to mitigate performance degradation.

\subsection{Challenges and Opportunities}

\begin{figure}[ht]
    \centering
    \begin{minipage}[t]{0.49\linewidth}
        \includegraphics[width=\textwidth]{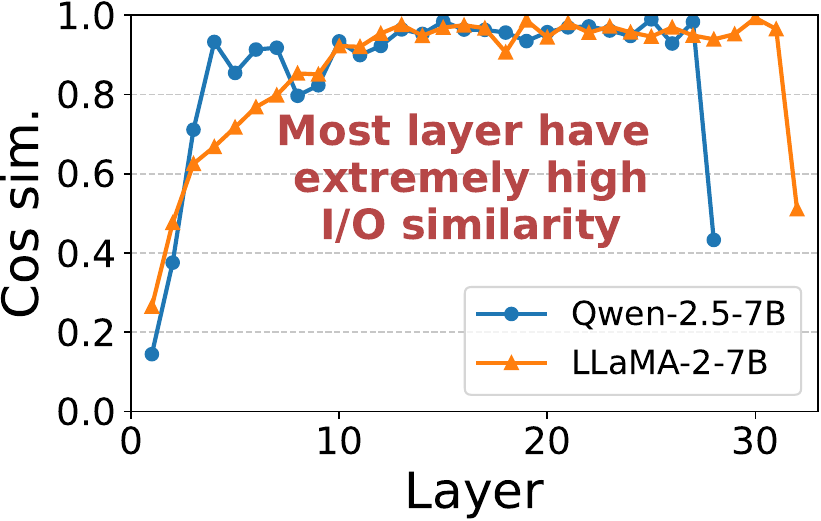}
    \end{minipage}
    \hfill
    \begin{minipage}[t]{0.49\linewidth}
        \includegraphics[width=\textwidth]{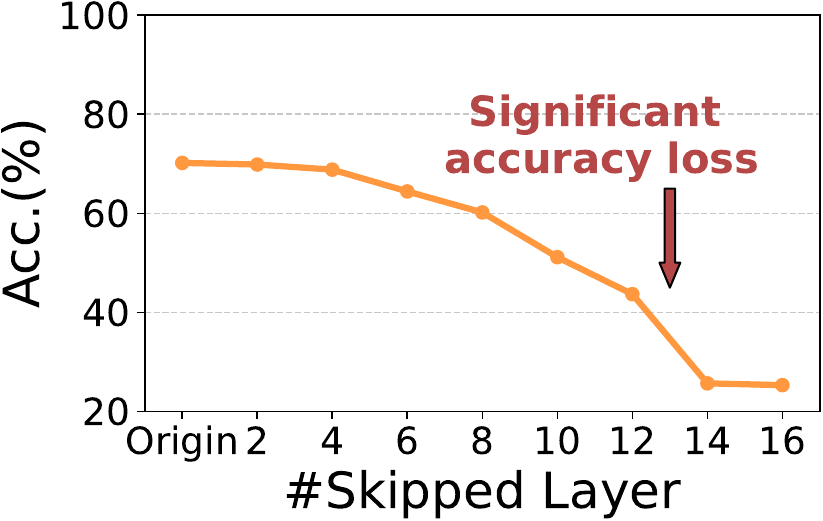}
    \end{minipage}
    \caption{Cosine Similarity and Model Accuracy Analysis. The left panel illustrates the cosine similarity across layers, indicating that representation similarity stabilizes after the initial layers despite early fluctuations. The right panel shows a precipitous decline in model accuracy as the number of skipped layers increases, emphasizing that minimal skipping maintains reasonable performance while excessive skipping results in a substantial accuracy drop.}
\label{fig:pro-cos and deg}
\vspace{-5pt}
\end{figure}

\textbf{Observation 1: Layer importance varies significantly across different models. }

We begin by analyzing the similarity between the Transformer layers' input and output representations to assess their relative importance. Specifically, given an input vector $X$ and an output vector $Y$, we measure their cosine similarity as:
\begin{align}
    \operatorname{Similarity}(\vec{x}, \vec{y})&=\frac{\vec{x} \cdot \vec{y}}{\|\vec{x}\|\|\vec{y}\|} \\
    &=\frac{\sum_{i=1}^n x_i y_i}{\sqrt{\sum_{i=1}^n x_i^2} \sqrt{\sum_{i=1}^n y_i^2}}
\end{align}

We interpret layers with high input-output similarity as less important, since the output remains close to the input, implying limited transformation and contribution during inference. Conversely, layers with lower similarity contribute more substantially and thus hold higher importance.

We evaluate the input-output similarity across layers of two distinct models on the CNN/DM reasoning task. As shown in Figure~\ref{fig:pro-cos and deg}, except for the initial and final layers, most intermediate layers exhibit high similarity, suggesting these layers can be skipped with minimal impact on performance. Notably, the similarity distributions differ markedly between models, underscoring the necessity of adaptive strategies tailored to the unique characteristics of each model.

\begin{figure*}[htbp!]
    \setlength{\abovecaptionskip}{3pt}
    \setlength{\belowcaptionskip}{3pt}
    \centering
    \includegraphics[width=0.9 \linewidth]{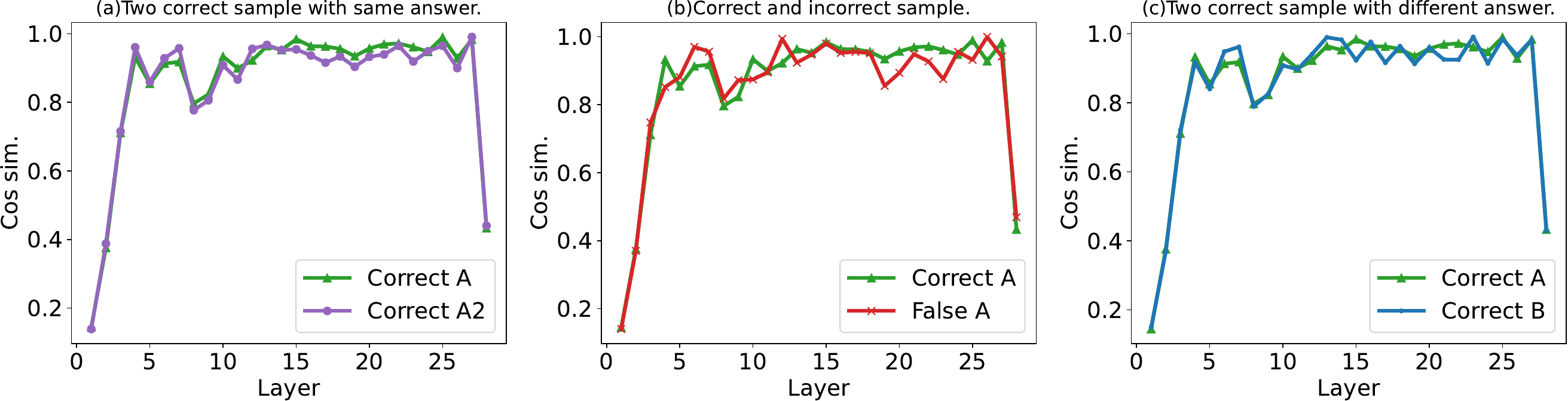}
    \caption{IO similarities between different samples on Qwen model with MMLU dataset.}
    \label{fig:pro-sample cos}
\vspace{-10pt}
\end{figure*}






\textbf{Observation 2: Static layer skipping leads to irreversible accuracy degradation.}

We conduct layer skipping based on inter-layer similarity for the Qwen-2.5-7B model by sequentially skipping layers with the highest similarity and tracking the corresponding accuracy changes (Figure~\ref{fig:pro-cos and deg}). The results reveal a steep accuracy decline as more layers are skipped. Specifically, for the 28-layer Qwen model, omitting just 6 to 8 of the most similar layers causes accuracy to deteriorate to the level of a random baseline. This demonstrates that simple static skipping cannot achieve a favorable trade-off between compression and performance retention.

To better understand the interplay between layer skipping and input variability, we analyze token-level input-output similarity over a small validation subset, focusing on a single-token generation task. Figure~\ref{fig:pro-sample cos} contrasts three output scenarios: identical correct outputs, outputs differing where one is correct and the other incorrect, and both correct but distinct outputs. The results highlight significant variation in similarity scores across tokens and layers, indicating that layer importance depends heavily on input content. Consequently, static similarity metrics aggregated over the entire dataset fail to capture this dynamic behavior, risking the omission of critical layers for specific inputs and resulting in accuracy loss.

These findings motivate the need for an input-adaptive, dynamic layer skipping mechanism that balances acceleration gains with accuracy preservation.

    
    
    
    
    



\textbf{Observation 3: Embedding changes are slow across layers}
We further analyze the input-output variations between consecutive layers for a fixed input and observe that the cosine similarity between embeddings or activations of adjacent layers remains consistently high across multiple models. Taking Qwen-2.5-7B as an example (Figure~\ref{fig:pro-cos and deg}), from the fifth layer onward, the similarity between consecutive layers exceeds 0.9, indicating that the embedding direction evolves gradually as the input propagates through the network. Notably, the most significant change occurs at the first layer. This finding suggests that such slow variation in embeddings can be exploited to design cooperative acceleration strategies, potentially mitigating the latency overhead introduced by dynamic layer-skipping mechanisms.



\subsection{Motivation}

Transformer-based models often exhibit significant inter-layer redundancy. Recent works such as SkipDecode and AdaSkip demonstrate that substantial portions of computation can be skipped without severe performance degradation. However, existing layer-skipping methods still face challenges in balancing skipping rates and prediction quality, largely due to input variability and architectural differences---even among models designed for the same tasks.

This challenge is particularly pronounced under dynamic input conditions. While static layer skipping’s limitations are well-known, compensating for information loss during runtime skipping remains underexplored. Our analysis of input-output similarity across Transformer layers reveals that many layers contribute minimal incremental information and can be safely skipped. Crucially, which layers can be skipped varies dynamically with input characteristics.

Motivated by this, we model dynamic layer skipping as a sequential decision-making process inspired by Markov decision processes. We propose the \textbf{DASH} framework---an adaptive system that dynamically identifies optimal computational paths based on real-time input analysis. By integrating a refined layer-skipping decision mechanism with an accuracy compensation strategy, \textbf{DASH} achieves a favorable trade-off between efficiency and performance. Furthermore, through asynchronous execution, our framework maximizes skipping while avoiding additional runtime overhead, making it well-suited for resource-constrained deployment.



\section{Methodology}

\begin{figure}[t!] 
\setlength{\abovecaptionskip}{3pt}
\setlength{\belowcaptionskip}{3pt}
\centering
\includegraphics[width=0.8 \linewidth]{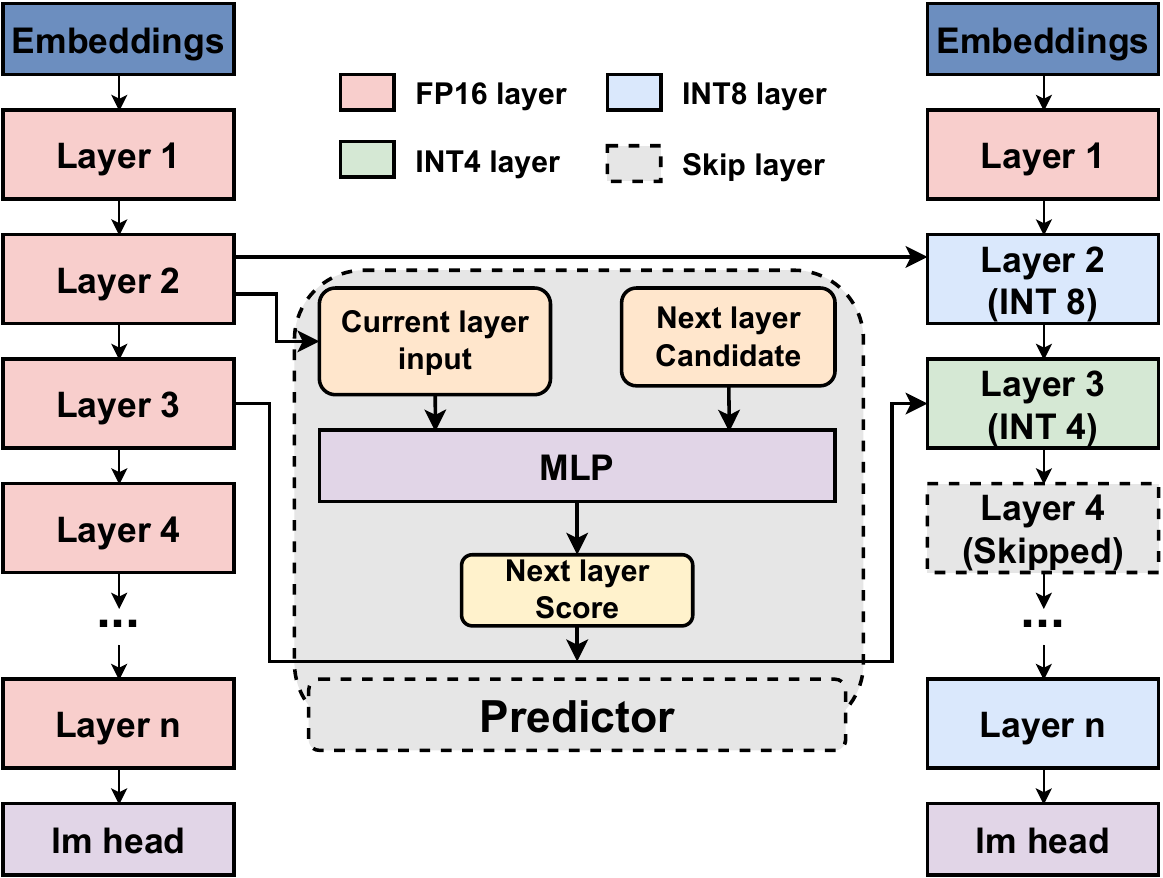}
\caption{Overview of the DASH Framework. This method first processes the embedding layer and maintains full-precision computation in the first Transformer layer. Starting from the second layer, the scoring model evaluates the next layer's state using the modified input of the current layer, dynamically selecting the next layer's state. When a layer is skipped, a compensation mechanism is activated based on the scoring results, effectively balancing inference speed and model accuracy.}
\label{fig:overview}
\vspace{-8pt}
\end{figure}

In this section, we present DASH, an adaptive dynamic layer-skipping framework that identifies optimal computational paths based on input characteristics. As illustrated in Figure~\ref{fig:overview}, DASH utilizes a scoring model to evaluate the importance of each Transformer layer during inference and dynamically decides whether to execute or skip the subsequent layer. To prevent accuracy degradation caused by aggressive skipping, DASH enforces strict constraints on skipping decisions and integrates a mixed-precision compensation mechanism, ensuring minimal impact on the model's original performance. Furthermore, to mitigate the additional latency introduced by the decision-making process, DASH exploits the high similarity between activation inputs of adjacent layers, enabling asynchronous evaluation that reduces overhead. Collectively, these components enable DASH to effectively reduce redundant computation while preserving accuracy, making it highly suitable for deployment in resource-constrained environments.

\subsection{Problem Formulation}

Consider a pre-trained LLM $M$ with $L$ Transformer layers.  Given an input sequence $X$, the standard forward pass computes all layers sequentially to produce output $f_M(X)$. 
Our objective is to reduce inference latency by selectively skipping certain layers while maintaining output quality. 
We formalize this as a layer-skipping decision problem: construct a binary decision sequence $S = {s_1, s_2, \dots, s_L}$, where each $s_i \in {0,1}$ indicates whether layer $i$ is skipped ($0$) or executed ($1$).  The subset of executed layers forms a computational path $P = \{l_{i}| s_i=1\}$.

Let $f_M(X)$ denote the model output when only layers selected by $S$ are computed. We aim to minimize the expected number of executed layers, $|S|=\sum_i s_i$, subject to an acceptable accuracy loss measured by a distance function $d(\cdot,\cdot)$ between full and partial outputs:
\begin{equation}
    \mathop{\min}_{D}\mathbb{E}_{X} [|S_{X}|] \ s.t. \ \mathbb{E}_X [d(f_{M}(X), f_{M,S_{X}}(X)] \leq \epsilon
\end{equation}

where $\epsilon$ is a user-specified tolerance threshold.

To capture dependencies between layers and adapt skipping decisions dynamically based on intermediate representations, we model the skipping strategy as a sequential decision-making process. Specifically, a policy $\pi_{\theta}$ parametrized by $\theta$ decides the next layer's execution state conditioned on the current layer's state $s_i$ and hidden features $h_i$:
\begin{equation}
    s_{i+1} \sim \pi_{\theta}(s_{i+1}|s_i ,h_i)
\end{equation}

The goal is to learn parameters $\theta$ that optimize the trade-off between computational cost and prediction quality. Formally,
\begin{equation}
    \theta^{*} = arg\mathop{\min}_{D} \mathbb{E}_{X}[\sum_{i=1}^{L} s_{i}] \ s.t. \ \mathcal{L}(\pi_{\theta}) + \mathcal{L}(M) \leq \epsilon
\label{eq:final theta}
\end{equation}
where $\mathcal{L}({\cdot})$ denotes the downstream task loss or overall model loss, incorporating both prediction accuracy and skipping regularization.

\subsection{Compensation Mechanisms Based on Differential Rewards}
Although DASH achieves significant inference acceleration by skipping a substantial number of layers, we observe that excessive skipping may cause non-negligible accuracy degradation. To address this issue, DASH incorporates a compensation mechanism based on differential rewards, which helps preserve model performance while enabling more aggressive layer skipping.

Specifically, when the scoring model decides to skip a layer $l_{i}$,  it implies that the input and output of the layer are sufficiently similar. However, instead of directly using the output of the previous layer $l_{i-1}$ as the input to the next layer $l_{i+1}$, we introduce computational compensation to approximate the skipped transformation.

DASH supports three actions for each layer: full execution, complete skipping, and partial computation. 
\begin{itemize}
    \item Full Execution performs the standard FP16 computation.
    \item Complete Skipping uses a scaling factor to approximate the transformation. The output is estimated as:
    \begin{equation}
    Y_{jt}^{i} = scale_j \cdot X_{jt}^{i}
\end{equation}
where the scaling factor is computed offline using a calibration dataset:
\begin{equation}
    scale_i = \frac{\sum_{j=1}^{N} \sum_{t=1}^{|T_j|} \frac{||Y_{jt}^{i}||}{||X_{jt}^{i}||}}{\sum_{j=1}^{N} |T_j|}
\end{equation}
Here, $X_{jt}^{i}$ and $Y_{jt}^{i}$ denote the input and output of layer $i$ for toke $t$ in sequence $j$, respectively.
    \item Partial Computation indicates that the layer’s computation is not essential in full precision. In this case, DASH uses low-bit quantization (e.g., INT4 or INT8) instead of FP16 to execute the layer with reduced cost while preserving critical information.
\end{itemize}
To unify this design, each layer’s execution state is encoded using a discrete value. Specifically, a state of \textbf{0} indicates that the layer is skipped and its output is approximated via scaling compensation. A state of \textbf{1} denotes that the layer is executed with low-precision INT4 computation, while \textbf{2} represents execution with INT8. Finally, a state of \textbf{4} corresponds to full-precision FP16 execution. This unified encoding enables flexible trade-offs between computational cost and accuracy on a per-layer basis.

The overall optimization objective remains unchanged (Equation~\ref{eq:final theta}): minimize the expected total computation cost while keeping accuracy loss within a predefined threshold.
Finally, DASH integrates these options into a unified decision process by assigning differential rewards based on the scoring model's output. The more computationally efficient the chosen action is, the higher the reward—conditioned on the predicted impact on accuracy. This differential reward guides the policy model to make fine-grained, input-adaptive decisions and enables DASH to explore more aggressive yet accurate skipping strategies.

\subsection{Layer Skipping Decision Mechanism}

Given the current layer $l_i$ of the model, our objective is to determine the execution state of the next layer $s_{i+1}$, where $s \in {0, 1, 2, 4}$. For accuracy preservation, we never skip the first layer and always set $s_1 = 4$. The final layer is also excluded from skipping decisions to ensure the integrity of output representations.


We employ a scoring model $G_{\theta}$ to implement the skipping policy. The score is computed based on the current hidden state, the current and next layer indices, and the current and candidate next layer states:

\begin{equation}
    g = G_{\theta} (h,l_i, l_{i+1}, s_i, s_{i+1})
\end{equation}

The scoring function $G_{\theta}$ is defined as:
\begin{equation}
    G_{\theta} = MLP([h_i;E(l_i);E(l_{i+1})]) - \alpha (s_{i+1} - s_i)
\end{equation}

where $E(\cdot)$ denotes the layer position embedding function and $[\cdot; \cdot]$ indicates vector concatenation. The term $-\alpha (s_{i+1} - s_i)$ serves as a regularization term that penalizes abrupt changes in layer execution states, encouraging smoother transitions and greater skipping continuity.

The MLP is a two-layer feedforward network:

\begin{equation}
    MLP(x) = W_3 \cdot GELU (W_2 \cdot GELU (W_1 \cdot x))
\end{equation}

where $W_1 \in \mathbb{R}^{(d_h + 2d_l) \times d_1}$, $W_2 \in \mathbb{R}^{d_1 \times d_2}$, and $W_3 \in \mathbb{R}^{d_2 \times 1}$, with $d_h$ as the hidden state dimension, $d_l$ as the layer embedding dimension, and $d_1$, $d_2$ as intermediate MLP dimensions.

To balance exploration and exploitation during training, we apply a temperature-based sampling strategy:

\begin{equation}
    P(s_{i+1} = s|h_{i}, s_{i}) \propto exp(\frac{G_{\theta}}{\tau_0 \cdot e^{-\alpha t}})
\end{equation}

where $\tau = \tau_0 \cdot e^{-\alpha t}$ is a temperature parameter that decays over training steps $t$ to reduce randomness over time.

\textbf{Differential Reward Function.}
To guide training, we introduce a location-sensitive differential reward, combining accuracy and efficiency:
\begin{equation}
    r_i = r_{acc} \cdot \omega_i + r_{eff}
\end{equation}

Here, $r_{\mathrm{acc}}$ measures the prediction accuracy, $\omega_i$ is a position-based importance weight, and $r_{\mathrm{eff}}$ rewards computational savings.

For classification tasks, $r_{\mathrm{acc}} = 1$ if the prediction is correct and $-1$ otherwise. For generation tasks evaluated with perplexity, the reward is calculated as:

\begin{equation}
    r_{acc} = \frac{\epsilon -|ppl_{f(x)} + ppl_{f_{s}(x)}|}{ppl_{f(x)}}
\end{equation}

The location importance weight $\omega_i$ takes both layer depth and skipping status into account:
\begin{equation}
    \omega_i = \frac{|S_{X}|-|S_{X,l}|}{|S_{X}|} \cdot \sigma(s_i -\frac{|S_{M}|}{|S_{X}|})
\end{equation}
where $|S_X|$ is the total sum of all layer states, $|S_{X,l}|$ is the cumulative state sum from layer 1 to $l_i$, $|S_M|$ is the maximum total state value, and $\sigma$ is the sigmoid function. This encourages skipping in earlier layers while considering execution intensity.

The efficiency reward is computed as:
\begin{equation}
    r_{eff} = \beta \cdot (4 - s_{i+1})
\end{equation}
This term provides higher rewards for more aggressive skipping.

\textbf{Training Objective.}
We optimize the scoring model via policy gradient reinforcement learning, aiming to maximize expected rewards:
\begin{equation}
    \nabla_\theta \mathcal{L}_{\mathrm{RL}} \approx-\sum_{i=1}^L r_i \nabla_\theta \log \pi_\theta\left( 
  s_{i+1}|s_i ,h_i\right)
\end{equation}

We jointly train the scoring model and the main model using a combined loss:

\begin{equation}
    \mathcal{L}_{all} = \mathcal{L}_{CE} + \lambda \mathcal{L}_{RL}
\end{equation}

where $\mathcal{L}_{\mathrm{CE}}$ is the cross-entropy loss for the main task and $\lambda$ is a balancing factor that controls the trade-off between accuracy and efficiency.

\textbf{Inference Strategy.}
During inference, we adopt a greedy strategy, selecting the next layer state with the highest score:
\begin{equation}
    s_{i+1} = arg \mathop{\max}_{s\in\{0, 1, 2, 4\}} G_{\theta} (h,l_i, l_{i+1},s_{i},s)
\end{equation}
This allows DASH to dynamically determine the execution path across layers, enabling efficient and adaptive inference.

\subsection{Asynchronous Skipping Decision for Latency Hiding}
Although the DASH policy dynamically reduces FLOPs during inference, the layer-wise decision process introduces non-negligible latency. Specifically, the decision score of each layer must be computed before executing that layer, which adds serial dependency and offsets the speedup benefits from skipping.

To address this issue, we propose an asynchronous decision mechanism that hides this latency. Our key insight is based on the empirical observation that hidden states, especially in transformer-based models, evolve gradually across layers. Thus, we can approximate the current layer’s hidden state using the previous one.

Specifically, when computing layer $l_i$, we approximate $h_{i+1}$ with a transformed version of $h_i$, scaled by a compensation factor $scale_i$:
\begin{equation}
    h_{i+1}^{'} = scale_i \cdot h_i
\end{equation}

We then use $h_{i+1}^{'}$ to compute the decision score for the next layer in parallel:
\begin{equation}
    s_{i+1} = \arg\max_{s \in \{0, 1, 2, 4\}} G_\theta(h_{i+1}^{'}, l_i, l_{i+1}, s_i, s)
\end{equation}

In this way, the scoring process for layer $l_{i+1}$ overlaps with the computation of laye

\vspace{-2pt}
\section{Evaluation}

In this section, we present a comprehensive evaluation of DASH’s inference-time performance. We begin by detailing the experimental setup, including benchmarks, baselines, and configurations. Then, we analyze the experimental results to demonstrate the effectiveness of our approach.

\subsection{Experimental Settings}

\noindent\textbf{Benchmarks.}  
We evaluate DASH on two representative large language models: LLaMA-2 (referred to as \textbf{LLaMA})~\cite{touvron2023llama} and Qwen-2.5-7B-Instruct (referred to as \textbf{Qwen})~\cite{yang2025qwen3}. The evaluation covers both generation and classification tasks. For text generation, we use WikiText-2~\cite{merity2016pointer} and CNN/DM datasets~\cite{see2017get}. For classification, we select two commonly used benchmarks: MMLU~\cite{hendrycks2020measuring} and ARC~\cite{clark2018think}.

\noindent\textbf{Models.}  
We utilize pre-trained models from HuggingFace and TorchVision Model Zoos. All baseline results are measured using FP16 precision. The models are evaluated under the same settings for a fair comparison.

\noindent\textbf{Calibration.}  
To compute the scaling parameters required for DASH’s skip compensation, we randomly sample 128 inputs from each dataset to construct a calibration set.



\noindent\textbf{Baselines.}
We compare our proposed method against four strong baselines for accelerating large language model inference: Early-Exit, SkipDecode, Adaskip, and RandomSkip. These methods represent diverse strategies for reducing computational costs during the generation process.

\textbf{1) Early-Exit}~\cite{fan2024not} enables the model to produce outputs from intermediate hidden states without completing a full forward pass. It dynamically selects exit points based on input complexity, thereby reducing inference overhead. This method integrates early-exit modules at specific transformer layers, allowing faster token generation with minimal quality degradation.
\textbf{2) SkipDecode}~\cite{del2023skipdecode} is a token-level early-exit strategy optimized for batch inference and key-value caching. Unlike traditional early-exit methods, it determines a unified exit layer per token across the batch at each sequence position. It enforces a monotonic decrease in exit layers across time steps, avoiding redundant key-value recomputation for past tokens. Rather than halting computation, it strategically skips lower and middle layers, focusing computation on higher layers.
\textbf{3) Adaskip}~\cite{he2025adaskip} is an adaptive sublayer-skipping method tailored for long-context inference. It identifies less informative layers based on similarity metrics and enables sublayer-level skipping to enhance efficiency without significant performance loss.
\textbf{4) RandomSkip} serves as a control baseline that randomly skips layers during inference according to a predefined probability distribution, without using any learned policy.

\subsection{End-to-End Result}

\begin{table*}[htbp]
\centering
\resizebox{\textwidth}{!}{%
\begin{tabular}{l||cccccccccccc}
\toprule
\multirow{3}{*}{} & \multicolumn{12}{c}{Qwen}                                                                                                 \\ \cline{2-13} 
                  & \multicolumn{3}{c|}{Wikitext2(ppl)}                                                 & \multicolumn{3}{c|}{CNN/DM(Rouge-L)}                                                & \multicolumn{3}{c|}{MMLU(Acc)}                                                      & \multicolumn{3}{c}{ARC-C(Acc.)}                                \\ \cline{2-13} 
                  & \multicolumn{1}{c|}{1.33x} & \multicolumn{1}{c|}{1.67x} & \multicolumn{1}{c|}{2.0x} & \multicolumn{1}{c|}{1.33x} & \multicolumn{1}{c|}{1.67x} & \multicolumn{1}{c|}{2.0x} & \multicolumn{1}{c|}{1.33x} & \multicolumn{1}{c|}{1.67x} & \multicolumn{1}{c|}{2.0x} & \multicolumn{1}{c|}{1.33x} & \multicolumn{1}{c|}{1.67x} & 2.0x \\ \hline
FP16              & \multicolumn{3}{c|}{6.62}                                                           & \multicolumn{3}{c|}{22.8}                                                           & \multicolumn{3}{c|}{70.2}                                                           & \multicolumn{3}{c}{60.6}                                       \\ \hline
Early-Exit        & \multicolumn{1}{c|}{13.1}  & \multicolumn{1}{c|}{35.4}  & \multicolumn{1}{c|}{1e3}  & \multicolumn{1}{c|}{21.6}  & \multicolumn{1}{c|}{18.4}  & \multicolumn{1}{c|}{10.7} & \multicolumn{1}{c|}{56.4}  & \multicolumn{1}{c|}{38.8}  & \multicolumn{1}{c|}{26.2} & \multicolumn{1}{c|}{45.5}  & \multicolumn{1}{c|}{38.5}  & 30.2 \\ \hline
RandomSkip        & \multicolumn{1}{c|}{85.2}  & \multicolumn{1}{c|}{173.4} & \multicolumn{1}{c|}{4e3}  & \multicolumn{1}{c|}{13.7}  & \multicolumn{1}{c|}{9.5}   & \multicolumn{1}{c|}{8.2}  & \multicolumn{1}{c|}{42.1}  & \multicolumn{1}{c|}{24.8}  & \multicolumn{1}{c|}{25.1} & \multicolumn{1}{c|}{36.8}  & \multicolumn{1}{c|}{30.6}  & 24.5 \\ \hline
SkipDecode        & \multicolumn{1}{c|}{36.8}  & \multicolumn{1}{c|}{96.5}  & \multicolumn{1}{c|}{1e3}  & \multicolumn{1}{c|}{20.2}  & \multicolumn{1}{c|}{13.5}  & \multicolumn{1}{c|}{8.1}  & \multicolumn{1}{c|}{63.5}  & \multicolumn{1}{c|}{60.4}  & \multicolumn{1}{c|}{55.3} & \multicolumn{1}{c|}{55.3}  & \multicolumn{1}{c|}{40.9}  & 33.5 \\ \hline
Adaskip           & \multicolumn{1}{c|}{8.12}  & \multicolumn{1}{c|}{18.4}  & \multicolumn{1}{c|}{77.4} & \multicolumn{1}{c|}{21.4}  & \multicolumn{1}{c|}{20.8}  & \multicolumn{1}{c|}{15.7} & \multicolumn{1}{c|}{66.8}  & \multicolumn{1}{c|}{58.1}  & \multicolumn{1}{c|}{43.2} & \multicolumn{1}{c|}{50.4}  & \multicolumn{1}{c|}{39.8}  & 31.1 \\ \hline
Ours              & \multicolumn{1}{c|}{7.43}  & \multicolumn{1}{c|}{19.9}  & \multicolumn{1}{c|}{53.2} & \multicolumn{1}{c|}{22.5}  & \multicolumn{1}{c|}{22.1}  & \multicolumn{1}{c|}{19.2} & \multicolumn{1}{c|}{69.7}  & \multicolumn{1}{c|}{67.5}  & \multicolumn{1}{c|}{61.0} & \multicolumn{1}{c|}{58.4}  & \multicolumn{1}{c|}{52.2}  & 36.5 \\ \bottomrule
\end{tabular}
}
\caption{Evaluation Results for Qwen model.
}

\label{tab:qwen-result}
\vspace{-3pt}
\end{table*}

\begin{table*}[htbp]
\centering
\resizebox{\textwidth}{!}{%
\begin{tabular}{l||cccccccccccc}
\toprule
\multirow{3}{*}{} & \multicolumn{12}{c}{LLaMA}                                                                                                                                                                                                                                                                                                        \\ \cline{2-13} 
                  & \multicolumn{3}{c|}{Wikitext2(ppl)}                                                  & \multicolumn{3}{c|}{CNN/DM(Rouge-L)}                                                & \multicolumn{3}{c|}{MMLU(Acc)}                                                      & \multicolumn{3}{c}{ARC-C(Acc.)}                                \\ \cline{2-13} 
                  & \multicolumn{1}{c|}{1.33x} & \multicolumn{1}{c|}{1.67x} & \multicolumn{1}{c|}{2.0x}  & \multicolumn{1}{c|}{1.33x} & \multicolumn{1}{c|}{1.67x} & \multicolumn{1}{c|}{2.0x} & \multicolumn{1}{c|}{1.33x} & \multicolumn{1}{c|}{1.67x} & \multicolumn{1}{c|}{2.0x} & \multicolumn{1}{c|}{1.33x} & \multicolumn{1}{c|}{1.67x} & 2.0x \\ \hline
FP16              & \multicolumn{3}{c|}{5.47}                                                            & \multicolumn{3}{c|}{27.7}                                                           & \multicolumn{3}{c|}{46.3}                                                           & \multicolumn{3}{c}{39.9}                                       \\ \hline
Early-Exit        & \multicolumn{1}{c|}{9.12}  & \multicolumn{1}{c|}{24.8}  & \multicolumn{1}{c|}{371}   & \multicolumn{1}{c|}{26.2}  & \multicolumn{1}{c|}{22.8}  & \multicolumn{1}{c|}{18.4} & \multicolumn{1}{c|}{40.6}  & \multicolumn{1}{c|}{34.9}  & \multicolumn{1}{c|}{32.5} & \multicolumn{1}{c|}{33.2}  & \multicolumn{1}{c|}{28.5}  & 26.3 \\ \hline
RandomSkip        & \multicolumn{1}{c|}{113.8} & \multicolumn{1}{c|}{1e3}   & \multicolumn{1}{c|}{3e3}   & \multicolumn{1}{c|}{15.8}  & \multicolumn{1}{c|}{8.4}   & \multicolumn{1}{c|}{8.1}  & \multicolumn{1}{c|}{33.8}  & \multicolumn{1}{c|}{26.1}  & \multicolumn{1}{c|}{25.1} & \multicolumn{1}{c|}{27.3}  & \multicolumn{1}{c|}{25.1}  & 24.9 \\ \hline
SkipDecode        & \multicolumn{1}{c|}{20.7}  & \multicolumn{1}{c|}{44.3}  & \multicolumn{1}{c|}{192.4} & \multicolumn{1}{c|}{22.8}  & \multicolumn{1}{c|}{19.6}  & \multicolumn{1}{c|}{15.0} & \multicolumn{1}{c|}{44.1}  & \multicolumn{1}{c|}{42.7}  & \multicolumn{1}{c|}{40.8} & \multicolumn{1}{c|}{37.1}  & \multicolumn{1}{c|}{33.7}  & 28.4 \\ \hline
Adaskip           & \multicolumn{1}{c|}{7.03}  & \multicolumn{1}{c|}{21.0}  & \multicolumn{1}{c|}{55.4}  & \multicolumn{1}{c|}{24.3}  & \multicolumn{1}{c|}{21.2}  & \multicolumn{1}{c|}{17.3} & \multicolumn{1}{c|}{45.9}  & \multicolumn{1}{c|}{42.8}  & \multicolumn{1}{c|}{32.4} & \multicolumn{1}{c|}{38.1}  & \multicolumn{1}{c|}{35.3}  & 29.7 \\ \hline
Ours              & \multicolumn{1}{c|}{6.61}  & \multicolumn{1}{c|}{13.6}  & \multicolumn{1}{c|}{38.8}  & \multicolumn{1}{c|}{26.8}  & \multicolumn{1}{c|}{23.3}  & \multicolumn{1}{c|}{20.1} & \multicolumn{1}{c|}{46.0}  & \multicolumn{1}{c|}{44.6}  & \multicolumn{1}{c|}{42.8} & \multicolumn{1}{c|}{38.6}  & \multicolumn{1}{c|}{36.4}  & 31.1 \\ \bottomrule
\end{tabular}
}
\caption{Evaluation Results for LLaMA model.
}
\label{tab:llama-result}
\vspace{-3pt}
\end{table*}

 Table~\ref{tab:qwen-result} and Table~\ref{tab:llama-result} summarize the performance of various layer skipping methods applied to Qwen and LLaMA models across multiple tasks, evaluated under acceleration ratios of 1.33×, 1.67×, and 2.0×.

As illustrated, our proposed method, DASH, consistently outperforms all baseline approaches at every acceleration level. Moreover, the performance gap widens as the acceleration ratio increases. At lower acceleration ratios, DASH incurs only negligible accuracy degradation—for instance, on the MMLU benchmark using the Qwen model, the accuracy drop is limited to 0.5\%. In contrast, when the acceleration ratio reaches 2.0×, competing methods suffer from severe performance degradation, nearing random guess levels. Conversely, DASH maintains robust performance, thereby achieving a significantly better trade-off between inference speedup and prediction quality.

\subsection{Ablation Study}

\begin{figure}[t!] 
\setlength{\abovecaptionskip}{3pt}
\setlength{\belowcaptionskip}{3pt}
\centering
\includegraphics[width=\linewidth]{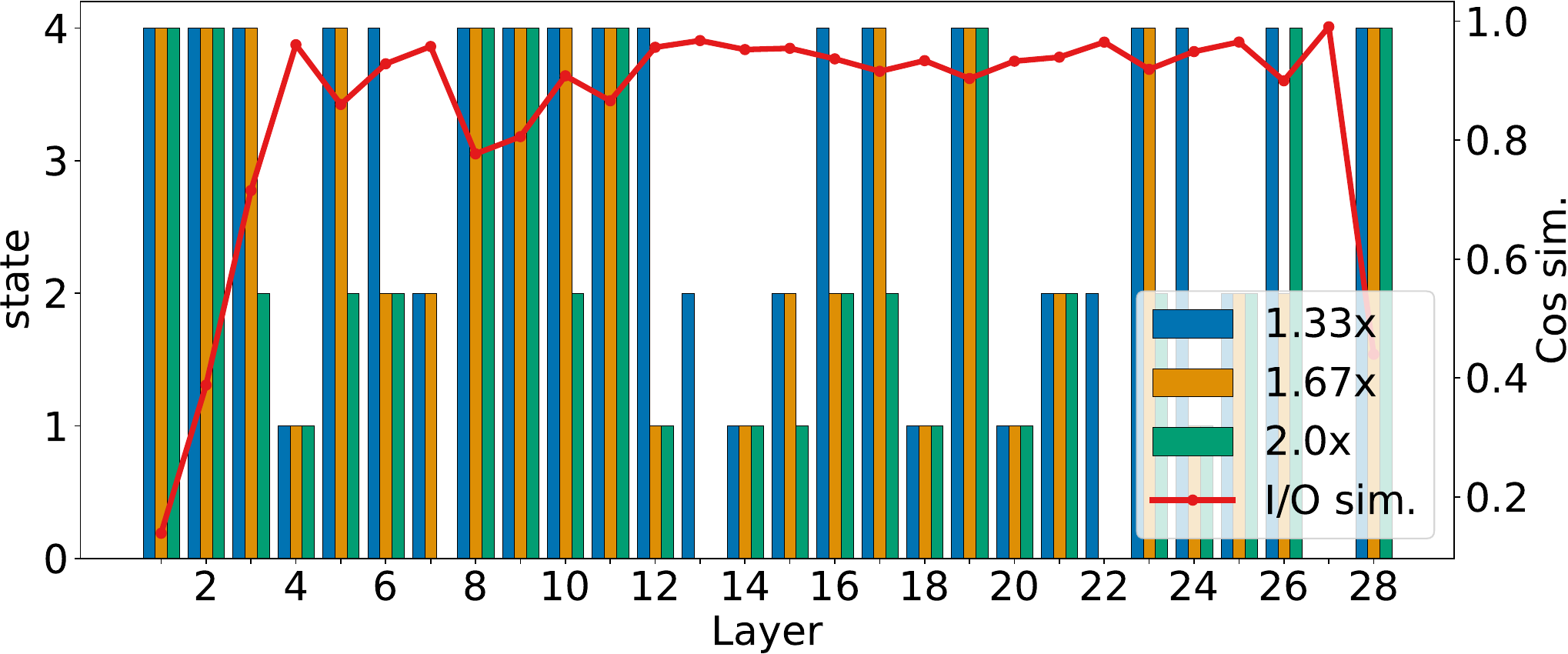}
\caption{I/O similarity and layer-skipping states at different speedup ratios.The higher the I/O similarity, the more aggressive the layer-skipping strategy becomes, preferentially selecting layer states with higher acceleration ratios.}
\label{fig:aba-different}
\vspace{-10pt}
\end{figure}

\textbf{Effectiveness of the scoring model.} 
We first validate the effectiveness of the decision model proposed in DASH. Using the Qwen model and the MMLU dataset, we analyze state function outputs at different acceleration ratios for the same sample, comparing them with the layer-wise input-output similarity computed for that sample. As shown in Figure~\ref{fig:aba-different}, although the specific skipped states vary across acceleration ratios, there is a clear positive correlation: layers exhibiting higher similarity tend to have a greater probability of being skipped, with an increased tendency towards complete skipping. This observation aligns well with our prior profiling analysis and demonstrates that DASH can accurately identify less critical layers, thereby achieving inference acceleration by selectively skipping layer computations.

\textbf{Effectiveness of the compensation mechanisms.}
Our dynamic layer-skipping strategy incorporates several compensation mechanisms, which enable flexible adaptation of skipping strategies and maximize the potential for layer skipping without significant accuracy degradation. We evaluate the impact of different compensation techniques on model accuracy, with results summarized in Table~\ref{tab:aba-compen}. Firstly, compared to static skipping methods, the introduction of a dynamic decision-making framework substantially restores accuracy, highlighting the efficacy of our adaptive approach. Secondly, the application of a simple scaling-based compensation method further improves accuracy relative to no compensation. Furthermore, integrating INT4/INT8 quantization within the layer-skipping pipeline also contributes to accuracy recovery. Collectively, these compensation mechanisms empower DASH to achieve superior accuracy-compression trade-offs compared to alternative approaches.

\begin{table}[]
\resizebox{\linewidth}{!}{
\begin{tabular}{|l|ccc|}
\hline
Method                         & \multicolumn{1}{c|}{1.33x} & \multicolumn{1}{c|}{1.67x} & 2.0x \\ \hline
Origin(Acc(\%))                 & \multicolumn{3}{c|}{70.18}                                     \\ \hline
1.Naive Static Skipping        & \multicolumn{1}{c|}{43.1}  & \multicolumn{1}{c|}{25.3}  & 25.2 \\ \hline
2.Dynamic Decision Skipping    & \multicolumn{1}{c|}{62.1}  & \multicolumn{1}{c|}{38.7}  & 24.9 \\ \hline
3.(2) + scale Compensation     & \multicolumn{1}{c|}{64.3}  & \multicolumn{1}{c|}{58.9}  & 42.7 \\ \hline
4.(3) + INT8 Compensation      & \multicolumn{1}{c|}{69.5}  & \multicolumn{1}{c|}{66.1}  & 59.2 \\ \hline
5.(3) + INT4/INT8 Compensation & \multicolumn{1}{c|}{69.7}  & \multicolumn{1}{c|}{67.5}  & 61.0 \\ \hline
\end{tabular}
}
\caption{Accuracy results during different compensation strategies.}
\label{tab:aba-compen}

\end{table}

\begin{figure}[t!] 
\setlength{\abovecaptionskip}{3pt}
\setlength{\belowcaptionskip}{3pt}
\centering
\includegraphics[width=0.85\linewidth]{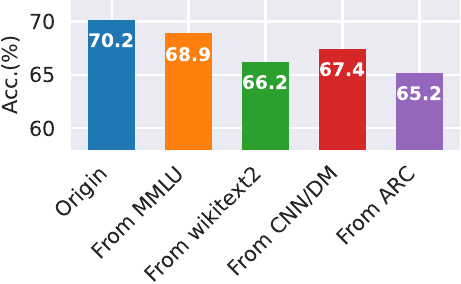}
\caption{Results on MMLU datasets. Decision system trained on different datasets by Qwen-2.5-7B}
\label{fig:aba-trans}
\end{figure}

\textbf{Transferability of scoring models.}
In practice, decision models are typically trained on task-specific datasets but are expected to generalize across diverse downstream tasks. To investigate this, we evaluate the performance of decision systems trained on different datasets when applied to the MMLU benchmark, with results presented in Figure~\ref{fig:aba-trans}. While all systems employ the same training methodology, their performance differs significantly due to variations in task-specific reward signals, where the decision model obtained using the corresponding training set performs the best. Nevertheless, even the least effective decision model achieves notable accuracy retention when executing dynamic layer skipping. This indicates a degree of transferability in our approach, although there remains substantial room for improving its generalization capability across tasks.

\section{Conclusion}
We identify dynamic layer skipping as a key approach to addressing the significant computational redundancy found in large Transformer-based language models. Our analysis reveals that static layer skipping methods often suffer from severe accuracy degradation due to their inability to adapt to input variability. To tackle this, we propose the DASH framework, which formulates layer skipping as a sequential decision-making process guided by a learned scoring model. Compared to existing static and heuristic skipping strategies, DASH consistently achieves higher acceleration ratios with minimal performance loss, thanks to its adaptive and input-aware skipping mechanism. Furthermore, the integration of compensation techniques effectively mitigates accuracy degradation, preserving model quality under aggressive skipping. We also demonstrate that the learned decision models exhibit promising transferability across different downstream tasks, indicating potential for broader applicability. 

\clearpage

\section*{Limitation}
Although we have demonstrated that DASH exhibits a certain degree of transferability across different tasks, the proposed scoring model still requires continuous updates to adapt to task requirements in long-running scenarios. Furthermore, due to the relatively low similarity in the initial layers of the model, the correction mechanism during asynchronous execution introduced in this solution may exhibit some instability. This instability tends to make the scoring model consistently avoid skipping the initial few layers, even when their I/O similarity has become sufficiently high, consequently preventing further acceleration.

\bibliography{ref}




\end{document}